\pdfoutput=1

\documentclass{article}

\usepackage[final,nonatbib]{neurips_2022} 

\usepackage{times}
\usepackage{latexsym}

\usepackage[T1]{fontenc}

\usepackage[utf8]{inputenc}

\usepackage{microtype}
\usepackage{cancel}
\usepackage{amsmath}
\usepackage{eqnarray}
\usepackage{cuted}
\usepackage{bbm}
\usepackage{flushend}
\usepackage{mathtools}   

\usepackage{microtype}
\usepackage{graphicx}
\usepackage{tabu}
\usepackage{multirow}
\usepackage{microtype}
\usepackage{arydshln}
\usepackage{lipsum}
\usepackage{booktabs} 
\usepackage{multicol,caption}
\usepackage{subcaption}
\usepackage[dvipsnames]{xcolor}

\title{
	Languages You Know Influence Those You Learn: Impact of Language Characteristics on Multi-Lingual Text-to-Text Transfer
}

\author{%
Benjamin Muller$^{*}$
\quad Deepanshu Gupta$^{\dag}$ 
\quad Siddharth Patwardhan$^{\dag}$ 
\quad Jean-Philippe Fauconnier$^{\dag}$\\
\bf{David Vandyke$^{\dag}$ 
\quad Sachin Agarwal$^{\dag}$}\\
$^{*}$INRIA, Paris, France\\
$^{\dag}$Apple, Cupertino, USA\\
{\tt benjamin.muller@inria.fr}\\
{\tt \{dkg,patwardhan.s,jfauconnier,dvandyke,sachin\_agarwal\}@apple.com}
}
\begin{document}

\date{}

\maketitle

\begin{abstract}
	In this work, we analyze a pre-trained mT5 to discover the attributes of cross-lingual connections learned by this model.
	Through a statistical interpretation framework over 90 language pairs across three tasks, we show that transfer performance can be 
	modeled by a few linguistic and data-derived features.
	These observations enable us to interpret cross-lingual understanding of the mT5 model.
	Through these observations, one can favorably choose the best source language for a task, and can anticipate its training data demands.
	A key finding of this work is that similarity of syntax, morphology and phonology are good predictors of cross-lingual transfer, significantly more than just the lexical similarity of languages.
	For a given language, we are able to predict zero-shot performance, that increases on a logarithmic scale with the number of few-shot target language data points.
\end{abstract}


\section{Introduction}
\label{sec:introduction}

Multi-lingual language models (LM), such as mBERT \cite{devlin-etal-2019-bert}, XLM-R \cite{conneau2019unsupervised}, mT5 \cite{xue2020mt5}, mBART \cite{liu-etal-2020-multilingual-denoising}, have been remarkably successful in enabling natural language tasks in low-resource languages through cross-lingual transfer from high-resource languages.
LM based pre-training and fine-tuning, combined with transfer learning resulted in state-of-art performance across various tasks \cite{pires-etal-2019-multilingual,libovicky2019language}.

In a typical cross-lingual transfer scenario, a single multi-lingual language model is pre-trained with large quantities of (unannotated) text from multiple languages.
It is then fine-tuned for a given natural language understanding task using human-labeled examples of that task in a {\em source} language.
Cross-lingual transfer occurs when this fine-tuned model can effectively perform this task on another language -- the {\em target} language -- without human-labeled data (called {\em zero-shot transfer}), or with only a few human-labeled examples in the target language (called {\em few-shot transfer}).

Recently, a line of work by \cite{hestness2017deep,kaplan2020scaling} has analyzed the scaling effects of parameters, corpus size and number of training steps on pre-training loss in language models \cite{devlin-etal-2019-bert}. 
\cite{hutter2021learning} extended this analysis to the out-of-distribution transfer setting and showed that the effective amount of data transferred from the training distribution to the target distribution follows a power law of the number of parameters and the amount of training data. 
Similarly, \cite{xia-etal-2020-predicting}  showed that the performance of a wide range of language tasks could be predicted with relatively good accuracy.
Their approach consists of parameterizing the experimental setting with both data-driven features, and linguistic features fed to a gradient-boosting model \cite{friedman2001greedy} to predict downstream performance.

Probing studies from \cite{pires-etal-2019-multilingual} and \cite{xue2020mt5} suggest that large multi-lingual language models exhibit zero-shot transfer ability and can deliver state-of-art performance for low-resource languages.
\cite{K2020Cross-Lingual} have suggested that ``structural similarity'' between the source and target languages is one of the most important factors regardless of the lexical overlap or word frequency similarity.  
Along similar lines, \cite{lauscher-etal-2020-zero} introduce a meta-regression framework and use it to predict cross-lingual task performance. \cite{lin-etal-2019-choosing} combined multiple features into a gradient-boosting model to predict zero-shot cross-lingual transfer performance. Finally, \cite{de-vries-etal-2022-make} combined multiple typological features in a single regression model to predict the cross-lingual transfer performance of XLM-R \cite{conneau2019unsupervised} in POS tagging. 
Our work extends their findings by presenting an interpretable statistical framework to explain  zero-shot and few-shot cross-lingual transfer. We do it across three tasks and 90 language pairs. While language similarity is a critical factor in effective cross-lingual transfer, we show that corpora size or language model performance in pre-trained models plays an equally important role. 

In our work, we try to better understand how multi-lingual pre-trained language models, such as mT5 \cite{xue2020mt5}, transfer {\em any} linguistic and semantic knowledge across languages.
There are no explicit cross-lingual signals provided to the model during pre-training.
Rather, unannotated texts from each language are presented to the model separately and independently of one another, and the model appears to implicitly learn cross-lingual connections.
The fact that this model exhibits cross-lingual transfer may suggest that it is somehow aligning its learned ``semantic spaces'' of different languages \cite{libovicky2019language,muller2021align}.
But, {\em are the cross-lingual connections between every language pair equally strong?}
{\em What properties of the source and the target language impact cross-lingual transfer performance?}
{\em Can we quantify the impact of those properties on the cross-lingual transfer?}
These are some of the key questions regarding effectiveness of cross-lingual transfer that naturally follow, and are the central theme of this work.

We posit that transfer between some languages is more dominant than others, based on the premise that not all language pairs are born equal \cite{wu-dredze-2020-languages}.
As highlighted by \cite{ruder2020beyondenglish}, designing an NLP system by mirroring what has been done on some high-resource languages (e.g., English) can lead to poor assumptions (e.g., ignoring the rich morphological connections between certain languages).
This approach is sub-optimal as it ignores specific properties of, say, Swahili or Arabic that could potentially see larger transfer benefits from ``non-traditional'' source languages.

Our contributions are three-fold:
	First, we establish an interpretable statistical framework to enable introspection into cross-lingual transfer in mT5.
	Next, using the above framework, we assess the impact of various factors on cross-lingual transfer.
	Finally, we derive linear connections between language similarity features, language model performance and number of target training samples for transfer learning.
A key finding of this work is that syntactic similarity, morphological similarity and phonological similarity are good predictors of cross-lingual transfer, significantly more so than just lexical similarity of language pairs.
For a given {\em \{source, target\}} language pair, we have the ability to predict zero-shot performance on the target language (for a given task), that is shown to increase on a logarithmic scale with the number of few-shot target language data points.



\section{Analysis Framework}
\label{sec:analysis-framework}

We start by first establishing an empirical framework to enable our analysis of cross-lingual transfer.
One of the things we want to understand is how the ``strength'' of cross-lingual transfer for a given language pair can be linked to a characteristic (or combination of characteristics) of that language pair.
In other words, can we ascertain if certain language pair characteristics will lead to a better cross-lingual transfer in mT5.
There is no universally agreed upon methodology for such a study, and a theoretical model of transfer learning across languages is not obvious.

For our analysis framework we draw inspiration from transfer learning literature \cite{mansour2009domain,pan2009survey,deng2013machine} as our starting point.
We analyze a pre-trained mT5 for pairs of languages ({\em source} language ($\mathcal{S}$) and {\em target} language ($\mathcal{T}$)) through observations of its performance ($S_{\mathcal{T}}$) in the target language on NLP tasks (e.g., NER, question answering, etc.), after fine-tuning it for the tasks using source language training data ($D_{\mathcal{S}}$), optionally fine-tuning with target language training data ($D_\mathcal{T}$), and evaluating the task on target language test data.
As such, cross lingual transfer can be captured through the function $f$:
\begin{equation}\label{eq1}
	S_{\mathcal{T}} = f(n_\mathcal{T}, n_\mathcal{S},  S_{\mathcal{S}}, D_{\mathcal{S}}, D_\mathcal{T}, \mathcal{A})
\end{equation}
\noindent where, ($S_{\mathcal{S}}$) is the performance on the source distribution, ($n_\mathcal{S}$) is the number of samples used for fine-tuning on the source distribution, ($n_\mathcal{T}$) is the number of samples used for fine-tuning on the target distribution, and ($\mathcal{A}$) is the learning algorithm with specific hyperparameter choices. 
Note that when $n_{\mathcal{T}}=0$, it is zero-shot transfer, while $0 < n_{\mathcal{T}} \ll n_{\mathcal{S}}$ it is few-shot transfer.

Since the data distribution for NLP tasks along with algorithmic complexity is hard to measure and observe, we make some reasonable simplifying assumptions to study cross-lingual transfer. We assume $n_\mathcal{S}$ to be constant and much larger than $n_{\mathcal{T}}$ (i.e. $n_{\mathcal{T}} \ll n_{\mathcal{S}}$), and we assume that we observe the performance of the model on the source distribution $S_{\mathcal{S}}$ so we discard $n_{\mathcal{S}}$ from equation~\ref{eq1}.
We utilize a measurable language similarity metric ($LS_{(\mathcal{T}, \mathcal{S})}$) and a language model performance metric ($LM$) instead of $D_{\mathcal{S}}, D_\mathcal{T}$ and $\mathcal{A}$ respectively in equation~\ref{eq1}.
With this, we can update our cross-lingual transfer equation~\ref{eq1} to:
\begin{equation}\label{eq2}
	S_{\mathcal{T}} = f(n_\mathcal{T}, S_{\mathcal{S}}, LS_{(\mathcal{T}, \mathcal{S})}, LM)
\end{equation}
where function $f$ captures the relationship between various language features and target language performance. 
If we are able to observe and measure the inputs and outputs of equation~\ref{eq2}, then this can  provide us with insights into the factors that enable cross-lingual transfer in mT5.
We search for an optimal combination of linguistic and/or data-driven features (the inputs to $f$) that can accurately estimate target language performance (the output from $f$) for {\em any} given language pair.
The implication is that features that are good predictors of target language performance are important for cross-lingual transfer.
In our experimentation framework, we consider several reasonable possibilities for $LS_{(\mathcal{T},\mathcal{S})}$ and modeling $LM$.

\subsection{Language Similarity}
\label{sec:lang_distance}

Similarity of languages ($LS_{(\mathcal{T},\mathcal{S})}$) can be assessed in many different ways.
A \textit{historical} linguistic approach \cite{hock2009principles} defines language relatedness through common parent or ancestor languages.
In a \textit{typological} approach \cite{delancey1983language}, language similarity is viewed through similarities in phonological, morphological and syntactic properties of languages.
Additionally, similarity between languages can be measured through \textit{statistical} means, using large corpora via tokens and character sequence overlaps.
Here, we are able to use some aspects of all of these similarity measures for model introspection.
We model language similarity through their lexical, morphological, phonological, and syntactic properties, which enables us to assess the impact of these on cross-lingual transfer.

\paragraph{Lexical:}

We define lexical language similarity by first computing the distribution of character n-grams for each source and target languages\footnote{We note that n-grams level lexical similarity is only a proxy of lexical similarity defined as the word-level vocabulary overlap. To avoid relying on imperfect tokenization, we define it using character level 3-grams.}.
To capture the similarities of the dataset that are involved in each experiment, we compute those distributions using the training dataset of each task.
We then compute a normalized Jensen-Shanon divergence ($JSD$) of the source distribution against the target distribution\footnote{Note that Jensen-Shanon divergence is a symmetric and smoothed version of Kullback–Leibler divergence.}. 
\begin{equation*}
	\textsc{Lex}_{(\mathcal{T}, \mathcal{S})}= 1-\frac{JSD(X_{\mathcal{S}}, X_{\mathcal{T}})}{max_{\mathcal{T},\mathcal{S}}JSD(X_{\mathcal{S}}, X_{\mathcal{T}})}
\end{equation*}
with $X_{L}$ defining the character 3-gram frequency distribution of the language $L$. We also define the vocabulary size ratio (\textsc{$V_r$}) between the source and target languages by dividing the vocabulary length of the target language by the vocabulary length of the source language. Finally, we measure the \textsc{SentLen} ratio by dividing the average sentence length of the target language with the average sentence length of the source language.  

\paragraph{Morphological:}

Following \cite{xia-etal-2020-predicting}, we use the Type-Token-Ratio (TTR) as a measure of how morphologically-rich a language is.
Based on this metric, we derive a Type-Token-Ratio similarity with:
\begin{equation*}
	\textsc{Morph}_{(\mathcal{T}, \mathcal{S})} = \frac{1}{K} \frac{TTR_\mathcal{T}}{TTR_\mathcal{S}}
\end{equation*}
where $K$, a normalization constant, is defined as $K = max_{(\mathcal{T}, \mathcal{S})}\frac{TTR_\mathcal{T}}{TTR_\mathcal{S}}$.

\paragraph{Phonological and Syntactic:}

We extract syntactic and phonological features from the World Atlas of Language Structures (WALS) database\footnote{The WALS is a  database of linguistic properties collected for a large number of languages \texttt{https://wals.info/}.
	We extract them using the lang2vec python package from \cite{littell-etal-2017-uriel}}.
For each type of property, we compute the intersection over union of the list of properties of the source language with the list of properties of the target language. We refer to those metrics as \textsc{Phono} for phonological similarity and as \textsc{Synt} for syntactic similarity.
For instance, for a given source language with the properties \textsc{Genitive-Noun-Order} and \textsc{Subject-Object-Verb Order} (e.g. Japanese) and a target language which has also the property \textsc{Genitive-Noun-Order} but a \textsc{Subject-Verb-Object Order} structure (e.g. French),  \textsc{Synt} would equal $\frac{1}{3}$ (1 shared property out of a union of 3 properties). 

\paragraph{Embedding Driven:}

Multilingual language models capture rich information about language similarities.
\cite{libovicky2019language} showed that language families can be retrieved from embedding representations computed with mBERT. \cite{muller2021align} showed that the centered kernel alignment (CKA) \cite{kornblith2019similarity} of the hidden representations across languages correlates strongly with downstream zero-shot cross-lingual performance. Finally, \cite{rama-etal-2020-probing} showed that a cosine-based embedding driven metric based on mBERT can be statistically explained by genetic and typological signals. 

Using those insights, we define an embedding-based language similarity.
(a) We compute a language centroid vector by average-pooling the hidden states of the mT5 encoder across a large sample of sentences in a given language.
(b) Then, based on these language centroid vectors, we compute a language similarity metric as:

\begin{equation*}
	\textsc{Emb}_{(\mathcal{T},\mathcal{S})} = cos(\bar{e_s}, \bar{e_t})
\end{equation*}
with $\bar{e_l} = \frac{1}{k}\sum_i e_{i,l}$ and $(e_{i,l})_{i}$, $k$ sentence embedding vectors in the language $l$.

\subsection{Language Modeling}
\label{sec:lm_def}

The $LM$ term in equation (\ref{eq2}) refers to performance of mT5 as a language model before any task-specific fine-tuning is done.
We define two metrics related to the pre-training mechanism and analyze their relationship with downstream task-specific performance. 
First, we adapt the language model score defined by \cite{salazar-etal-2020-masked} to the denoising objective used to pre-train T5.
We compute the output log-likelihood of the model on span-masked sentences.
Formally, for a collection of sentences $X={x_1, \ldots, x_{|X|}}$ with $x_s=(x_1, \ldots, x_{|x|})$ in the language $L$, we define the language model performance as:
\begin{equation*}
	LM_{\mathcal{L}(L)} = 
	\frac{1}{|X|}\sum_x\sum_{i\in s} log(p(x_i|x \setminus x_i))
\end{equation*}

For our second approach, we define an Exact-Match accuracy metric as follows:
\begin{equation*}
	LM_{EM(L)} = 
	\frac{1}{|X|}\sum_{x,s}\prod_{i\in s} \mathbbm{1} (\hat{x}_i=x_i)
\end{equation*}
with $\hat{x}_i$ is the greedy-decoded model prediction of a sequence $x$ after masking spans indexed by~$s$.
$LM_{EM(L)}$ simply captures how accurate the span predictions of the model are when feeding it masked sentences. We follow strictly the pre-training span masking procedure defined in \cite{xue2020mt5}. We estimate both these statistics on the training dataset of each task.


\section{Experiments}
\label{sec:experiments}

\begin{table*}[t!]
	\centering
	\small
	\scalebox{0.79}{ 
		\begin{tabular}{lrrrrrrrrrrr}
			\toprule
			Task &
			
			$LM_{\mathcal{L}(\mathcal{S})}$&
			$LM_{\mathcal{L}(\mathcal{T})}$&
			$LM_{EM(\mathcal{S})}$ &
			$LM_{EM(\mathcal{T})}$ &
			\textsc{Syntax} & \textsc{Phono} & \textsc{Morph} & \textsc{Lex} & \textsc{Emb} & \textsc{$V_r$} & \textsc{SentLen}  \\
			\midrule
			
			

			
			
			{\tiny QA} & 6.6 & 7.3* &-15.4* &-18.1*&38.4* &29.6* &-40.7* &19.3* &8.9* &-40.7* &3.4 \\ 
			{\tiny NER} & 22.1* &1.3 &1.3 &6.6&2.4 &14.7* &-2.4 &23.6* &16.2* &-2.4 &-27.1* \\ 
			{\tiny XNLI} &6.5 &23.9*& -26.5* &4.1 &-0.1 &0.1 &10.8* &-6.3 &1.3 &10.8* &-12.4* \\ 
			
			\bottomrule
			
		\end{tabular}
	}
	\caption{Pearson correlation between the features introduced and the cross-lingual transfer performance in the zero-shot setting - measured as $S_{\mathcal{T}}-S_{\mathcal{S}}$ - for XNLI, QA and NER across source languages ($\mathcal{S}$) and target languages ($\mathcal{T}$) (* indicates statistical significance).
	}
	\label{tab:correlation}
\end{table*}

Based our framework made of equation~\ref{eq2} and the features presented in the previous section, we now present our experiments. We start by presenting a bi-variate analysis between each feature and the cross-lingual transfer performance. Then, we present a meta-regression that combines the multivariate effect of the features on the cross-lingual transfer performance.

As mentioned before, we focus on the mT5 framework, a multi-lingual adaptation of T5 \cite{raffel2019exploring}. 
T5, \textit{Text-To-Text Transfer Transformer}, formulates any NLP tasks as sequence generation. 
If the task is a classification or regression, we generate the label token by token as if it were natural language. 
In a nutshell, the T5 framework abstracts away the output feature engineering from meaningless indexes to meaningful language tokens.
Its architecture is a Transformer \cite{vaswani2017attention} encoder-decoder, pre-trained with a span-masking objective closely inspired by the BERT model \cite{devlin-etal-2019-bert}. 
We run our cross-lingual analysis on the base version of mT5.

Our analysis is conducted on Arabic, Bengali, English, Finnish, Indonesian, Russian, Swahili, Spanish, German, Hindi.
Not all languages have training data for all the three tasks we work with but each task gets at least 7 languages.
We report the detailed list of the languages used for each task in the Appendix in Table~\ref{tab:data_langs}. 
Each language is used both as a source language ($\mathcal{S}$) and as a target language ($\mathcal{T}$) leading to up to 90 language pairs.\footnote{Having access to many more languages for XNLI than for the other tasks, we extend our zero-shot cross-lingual transfer experiment to 7 extra languages, namely (Bulgarian, Greek, French, Turkish, Urdu, Vietnamese, and Chinese).} 

Additionally, we focus on three tasks: Natural Language Inference (NLI), Name-Entity Recognition (NER), and Question Answering (QA). 
For NLI, we use the XNLI dataset \cite{conneau-etal-2018-xnli}, for NER the PANX dataset \cite{ghaddar-langlais-2017-winer} and for QA the TyDiQA (for Typologically Diverse Question-Answering) dataset \cite{clark-etal-2020-tydi}. 

We report the standard evaluation score for each task: for XNLI we report the accuracy, for NER the F1 score and for QA the exact-match of the predicted answers with the gold answers.
To allow comparison across languages, for each task, we control for the number of training samples in the source languages (100k samples for XNLI, 10k for NER, and 2.2k for QA for each source language). 
For the target language, we run experiments with $n \in \{0, 10, 30, 50, 100, 250\}$ for all three tasks and include $n \in \{500, 750, 1000\}$ for NER and XNLI.

\begin{table}[b]
	\centering
	\begin{tabular}{lrrr}
		\toprule
		Transformation of $n$ & QA & NER  & XNLI  \\
		\midrule
		$n$ & 67.3 & 58.1 & 14.2 \\
		$log(n)$ & \bf 83.9 & \bf 69.0 & 14.8 \\
		$log^2(n)$&  76.9 & 67.8 & \bf 18.0\\
		\bottomrule
	\end{tabular}
    \vspace{0.2cm}
	\caption{Pearson Correlation between $S_{\mathcal{T}}$ and various transformation of the number of samples in the target language $n$ used for fine-tuning mT5 for XNLI, QA and NER. (with $n\in\{0, 10, 30, 50, 100, 250, 500, 750, 1000\}$).}
	\label{tab:log_n}
\end{table}

\subsection{Bivariate Correlation Analysis}
\label{sec:bivariate_corr_analysis}
We start this study with bi-variate correlation analysis over predictors introduced in section~\ref{sec:lang_distance}.
We report in Table \ref{tab:correlation} the Pearson correlation of each predictors with cross-lingual transfer performance for each task. 
We indicate with $*$ statistically significant results.\footnote{We run a two-tailed statistical significance test on the Pearson correlation and consider correlation to be significant when the p-value $\leq5\%$}
Among others, we find that \textsc{Lex}, \textsc{Phono} and \textsc{Emb} have significant correlation for QA and NER.
For XNLI, we find that $LM_{\mathcal{L}(\mathcal{T})}$ and \textsc{Morph} are correlated significantly with cross-lingual transfer.  This suggests that those predictors are good candidates to be used in a predictive regression model.

Overall, the correlations have similar trends  across tasks (correlation signs are the same across tasks in most cases) but with significant differences in strengths. We observe a few key differences for some features.
For instance, the syntactic similarity correlates strongly with cross-lingual transfer for Question Answering but does not with NER and XNLI.
These results suggest that a multivariate linear model should be task-specific\footnote{We note that our results extend previous findings from \cite{pires-etal-2019-multilingual}.}.

Still, we note that despite the observed correlations, each of these metrics provides an incomplete view of the relationship between language distance and cross-lingual transfer. Indeed, the correlation is never close to 1 which means that none of these features alone is informative enough to predict cross-lingual transfer performance. 

For instance, character level 3-grams is only limited to languages that share the same script.
Languages with different script have very high lexical divergence. We note that the divergence is never undefined (infinite) even between languages written in different scripts.
This is due to the numbers and residual Latin tokens used in the Arabic, Russian and Korean datasets.

For instance, the divergence between Swahili and Arabic is very close to the one from Swahili to Hindi but Arabic leads to a much better transfer.
Additionally, in some cases our language similarity metrics fail to explain cross-lingual transfer.
For instance, for QA, the embedding-based similarity between Indonesian and Swahili is significantly higher than the one between Finnish and Swahili.
Still, the transfer to Swahili is better when the source is Finnish.

We run the same analysis with language model performance measured with log-likelihood and exact-match accuracy (cf. Section~\ref{sec:lm_def}).
Language model performance exhibits strong correlation, but surprisingly, the correlation is counter-intuitive.
For instance, for QA, the higher the language model accuracy on the target language the lower the cross-lingual transfer.

\begin{figure}[t]
	\centering
	\includegraphics[width=0.7\columnwidth]{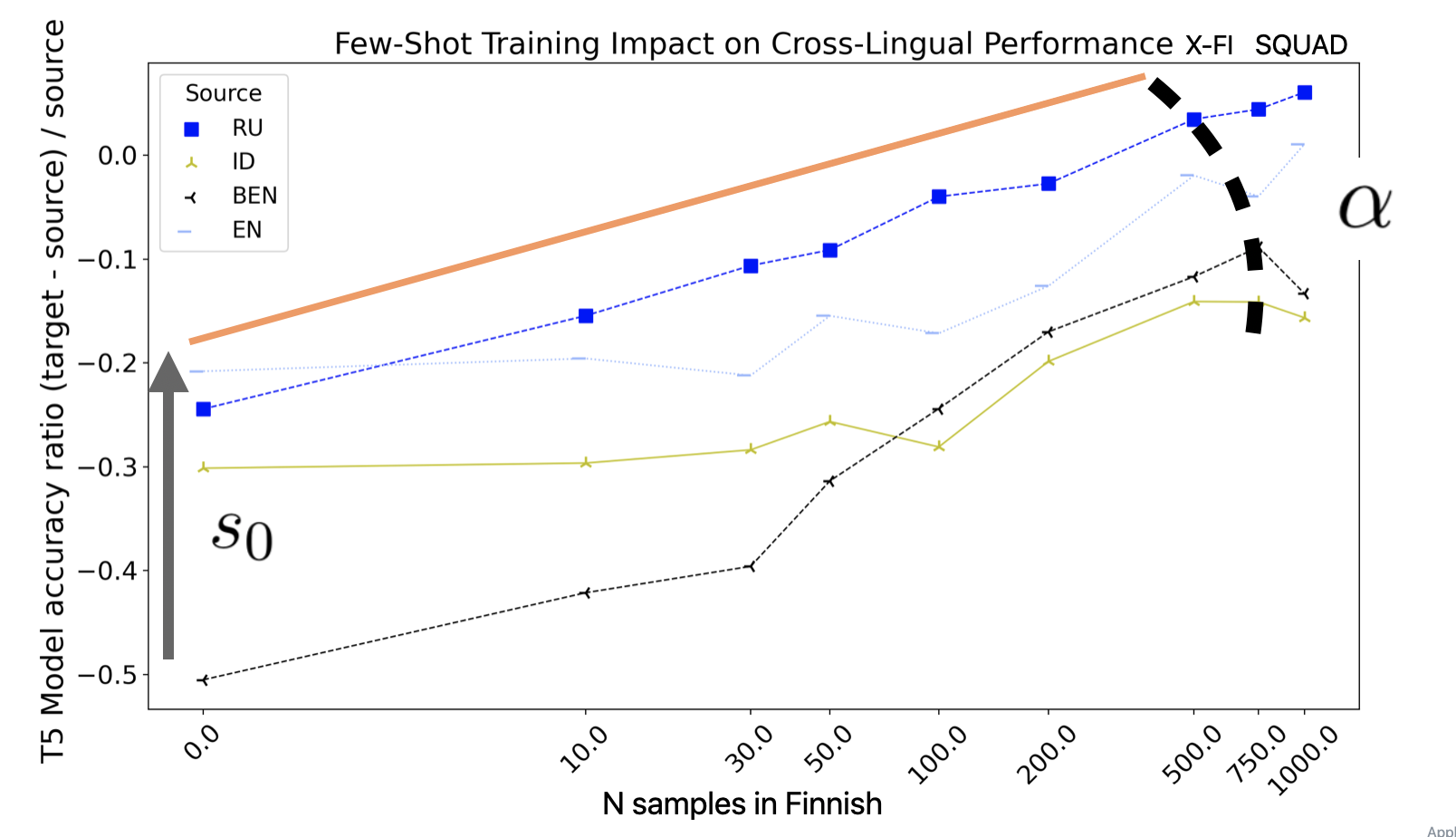}
	\caption{Zero-shot vs. N-shot Cross-Lingual Transfer for QA when transferring to Finnish from Russian, Indonesian, Bengali and English.}
	\label{fig:zero_shot_vs_few_shot}
\end{figure}

To gain more insights into $f$, from Equation~(\ref{eq2}) in Section~\ref{sec:analysis-framework}, we analyze the bivariate relationship between the target performance $S_{\mathcal{T}}$ and the number of samples $n$ in the target language.
As illustrated in Figure~\ref{fig:zero_shot_vs_few_shot} with the n-shot cross-lingual performance from various languages to Finnish for Question Answering, there is a strong linear relationship between the log of the number of samples in the target language and downstream accuracy.
We report in Table~\ref{tab:log_n}, the Pearson correlation between transformation of the number of samples with downstream performance in the target language for the three tasks.
We find that downstream $n$-shot cross-lingual transfer performance correlates strongly with $log(n)$ and the most for two of the three tasks. However, for XNLI, we note that $log^2(n)$ is a stronger predictor and that the absolute correlation is weaker than for the two other tasks. We explain this by the fact that the absolute performance of XNLI in the few-shot setting increases only moderately compared to the zero-shot setting.

In summary, we introduced several predictors that, overall, correlate linearly and strongly with cross-lingual performance.
In the following sections, we show that these predictors can be combined linearly to predict cross-lingual transfer reasonably well.


\subsection{Meta-Regression of Cross-Lingual Performance}

The bivariate analysis in the prior section principally motivates us to hypothesize that cross-lingual transfer has two components to itself.
A source-target similarity component that dominates the zero-shot learning of a target language.
While the logarithmic relation in number of target samples governs n-shot learning.

In consequence, we simplify $f$ as follows:
\begin{equation*}
	S_{\mathcal{T}} = S_\mathcal{T}(0)+\alpha~log(n+1)
\end{equation*}
\begin{equation}
	S_{\mathcal{T}} = s_0+\alpha~log(n+1)
\end{equation}
with:
\begin{equation*}
	~S_{\mathcal{T}}(0) = f_0(S_{\mathcal{S}},T, S, LM,\mathcal{T})
\end{equation*}
\begin{equation*}
	\alpha = f_1(S_{\mathcal{S}},T, S, LM,\mathcal{T})
\end{equation*}
Note that $S_\mathcal{T}(0)$ corresponds to the performance of the model in the zero-shot setting.
While $\alpha$ is a function of source-target language, algorithm and task difficulty that governs the slope of the performance curve.
As such, we will break our analysis into two separate components and study them as such.

\paragraph{Feature Selection:}
\label{sec:feature_selection}
While we study an array of diverse feature transformation and combination, however, we do filter out most of them.
To keep only the most simple and relevant features, we use a Lasso Regression \cite{tibshirani1996regression}.
We run feature selection with recursive feature elimination using the absolute value of each coefficient.
We start with all the features, we fit the regression and we iteratively remove the feature with the smallest coefficient.

\paragraph{Evaluation:}
Following \cite{xia-etal-2020-predicting} we fit our meta-regression model with $l$-folds cross-validation and define each fold so that they include only the observations for a single target language.
We fit the model on the concatenation of $l-1$ language-folds and we evaluate on the $l^{\rm{th}}$ fold.
We report the average $l$-cross-validation score, which corresponds to the average computed on all the target languages.
We evaluate the performance of the regression using the Root Mean Squared Error (RMSE), a standard metric to evaluate a regression model.

Additionally, for a given language and a number of annotated samples, we predict which source language leads to the best performance according to our regression.
We do by solving argmax over all possible language pair for the Equation \ref{eq2}. 
We then measure the accuracy of the prediction by comparing it to the actual best source language.
We denote this accuracy metric $A_{src}$.

\subsubsection{Zero-Shot Transfer Meta-Regression}

We find that for the three tasks, the zero-shot cross-lingual transfer can be effectively modeled with a linear combination of afore mentioned features.
We summarize in Table~\ref{tab:regression_results_zero_shot} the performance of the regression.
Note, a simple model with most relevant features is able to explain a lot of the variance in the QA performance for zero-shot transfer with RMSE as low as 5.48.
Additionally, our model can be used to predict the best source language with a relatively good accuracy.
For instance, on QA, the model can predict the correct source language in 64\% of the cases.

We present in Figure~\ref{fig:equation_linear} the final linear relationships between the cross-lingual performance and the selected predictors after recursive feature elimination for the three tasks. For interpretation purposes, we report the coefficients of the regression fitted on all the language-folds.
We find that few features such as syntactic, morphological and lexical similarity are strong predictors of transfer learning performance.
Surprisingly, despite the strong bivariate relationship, the language centroid similarity does not offer helpful signals in the presence of other features. Additionally, we find that language model performance as measured with the exact match accuracy is a useful predictor of task-specific performance.


\begin{table}[t]
	\centering
	\begin{subtable}[t]{0.50\linewidth}
		\small
		\centering
		\begin{tabular}{lcc}
			\toprule
			Task  & RMSE  & Top-1 Source \\
			&       &  Prediction Accuracy  \\
			\midrule
			QA &  5.48 & 64.3\%  \\
			NER &  9.80 & 50.0\%  \\
			XNLI &  4.26 & 16.4\%. \\
			\bottomrule
		\end{tabular}
		\caption{Zero-shot meta-regression fit; RMSE and $A_{src}$\\ top-1 source language prediction accuracy.}
		\label{tab:regression_results_zero_shot}
	\end{subtable}
	\begin{subtable}[t]{0.30\linewidth}
		\small
		\centering
		\begin{tabular}{lrr}
			\toprule
			Task & RMSE \\
			\midrule
			SQUAD &  5.69 &\\
			NER &  7.60 &\\
			XNLI  &  8.13 &\\
			\bottomrule
		\end{tabular}
	\caption{$n$-shot meta-regression RMSE of $f_1$ for each task.}
\label{tab:regression_results_few_shot}
\end{subtable}
\caption{Meta-regression results in zero-shot and $n$-shot settings.  The Root Mean-Square-Error (RMSE) corresponds to an average error of our regression in absolute points.}
\label{tab:x_shot_results}
\end{table}

For all the three tasks, the language model performance on the target language has a significant and positive coefficient.
This suggests that a better language modeling of the target language leads to better task-specific prediction.

\begin{figure}[t!]
	\begin{multline*}
		\mbox{\small (QA)}~~S_{\mathcal{T}}(0) =-65.38 + 0.62 ~S_{\mathcal{S}}+ 56.49~Syn+156.4~Phono-29.73~Morph+129.3~LM_{EM(\mathcal{T})}
	\end{multline*}
	\begin{multline*}
		\mbox{\small (NER)}~~S_{\mathcal{T}}(0) = -42.8+1.07~S_{\mathcal{S}}
		+14.63~Syn+68.22~LM_{EM(\mathcal{T})}+4.09~Lex
	\end{multline*}
	\begin{multline*}
		\mbox{\small (XNLI)}~~S_{\mathcal{T}}(0) =27.62+0.64~S_{\mathcal{S}}+ 10.12~Phono-1.2~Morph+46.87~LM_{EM(\mathcal{T})}+1.2~Lex
	\end{multline*}
	\caption{Regression models after feature selection for the three tasks to predict zero-shot cross-lingual transfer. All the coefficients are non-null (with a p-value $\leq 0.05$). All the variables are scaled between 0 and 1, allowing comparison across coefficients.}
	\label{fig:equation_linear}
\end{figure}

From the equations in Figure~\ref{fig:equation_linear}, we discover the following relationships between predictors and performance on the target language.
\begin{itemize}
	\item  For every 1\% improvement in language modeling accuracy on the target language, we can expect about 1.3\% improvement in target language performance for QA, 0.7\% for NER  and 0.5\% for XNLI.
	\item Every 10 points increase in syntactic similarity leads to a corresponding increase of around 5 points in downstream accuracy for QA, 1.4 for NER.
	\item For QA and XNLI, we find that downstream cross-lingual performance is strongly related to phonological similarity: 10 points increase in language similarity can result in up to 15.6 points increase in cross-lingual performance in QA and 10 points in XNLI.
	\item Surprisingly, morphology has large negative effect on cross-lingual transfer for QA and more moderate one for XNLI.
\end{itemize}

\subsubsection{N-Shot Cross-Lingual Meta-Regression}

For $n$-shot cross-lingual transfer performance prediction, we run the same feature selection for $f_1$.
We found that the two main factors that are reliable predictors for each QA, NER and XNLI task performance: (i)~number of samples available for training target language and the (ii)~the zero-shot performance.
We notice that the influence of the language similarity metrics and language model performance usually dies out once more target language samples are available for training. 
These results are consistent over all language pairs and tasks. 
We report the performance of the regression for the $n$-shot setting in Figure~\ref{tab:regression_results_few_shot}.

We enumerate here several observations from our experiments.
For QA, we notice that an increase in the number of data samples in the target language by a factor of 10 has a corresponding increase in target language performance by an average of over 5\%.
The slope of this increase, however, also depends on the zero-shot performance and language similarity.
Observe that higher language similarity equates to an increase in zero-shot performance, which in turn \underline{reduces} the influence of target language annotated data.
For XNLI, we learn that the slope due to number of samples is pretty low and consistent over languages.
The slope is very highly dependent on the few shot accuracy
For NER, we find that a 10x increase in target training data has on average a 7\% increase in the task performance.
The broad takeaway from these observations is that target language performance is heavily predictable by training samples for all tasks.




\section{Conclusion and Future Work}

In this work, we analyze a pre-trained mT5 to study the nature of cross-lingual transfer.
Through model interpretation experiments over multiple language pairs and tasks, we show that transfer can be modeled statistically by a few linguistic and data-derived features.
We show that syntax, morphology, and phonology of languages are good predictors of cross-lingual transfer (significantly better than lexical similarity of languages). 
We also demonstrate that language model performance is an informative predictor of cross-lingual performance, providing an off-the-shelf metric to inform cross-lingual transfer.

Since, reproducing these relations for all language pairs and all tasks is a huge drain of resources, a natural future work for this line of research is to generalize these findings and relations over all tasks instead of the 3 that we did. 
Similarly, we focused on mT5 only due to its simplicity of architecture, an interesting follow-up would be to derive more general relations for other architectures like mBART, mBERT etc.
Additionally, we would like to better understand influence of {\em multiple} source languages on each target language -- i.e., {\em is there a combination of several source languages that could enable better cross-lingual transfer?} 
Finally, we would like to leverage this knowledge to influence the mT5 pre-training methodology to better facilitate few and zero-shot transfer across tasks. 

\bibliography{acl2020,custom}
\bibliographystyle{alpha}

\clearpage

\appendix

\section{Reproducibility}
\label{sec:appendix}

\subsection{Fine-tuning and Evaluation Data}

For NER, we use the data from \cite{ghaddar-langlais-2017-winer}, for XNLI we use the data from  \cite{conneau-etal-2018-xnli} and for Question Answering (QA), we use the TyDiQA dataset from \cite{clark-etal-2020-tydi}.
All those datasets are downloaded and pre-processed using the scripts from the XTREME benchmark \cite{xtreme} available at \texttt{https://github.com/google-research/xtreme}.
We work with the languages listed in Table~\ref{tab:data_langs}.

\begin{table}[h]
	\footnotesize
	\centering
	\scalebox{1.0}{ 
		\begin{tabular}{lrrr}
			\toprule
			&  \multicolumn{3}{c}{\underline{Available Data Train/Test}} \\
			Languages (iso)&XNLI &  TyDiQA & NER \\ 
			\midrule
			Arabic (AR) & 1/1 & 1/1 &   1/1  \\
			Bengali (BN) & $\emptyset$  & 1/1 &  $\emptyset$  \\
			English (EN) & 1/1  & 1/1 & 1/1  \\
			Finnish (FI) & $\emptyset$  & 1/1 &   1/1 \\
			Indonesian (ID) & $\emptyset$  & 1/1 & 1/1  \\
			Russian (RU) & 1/1  & 1/1 & 1/1  \\
			Swahili (SW) & 1/1  & 1/1  &$\emptyset$/1   \\
			Spanish (ES) & 1/1  & $\emptyset$  & 1/1   \\
			German (DE) & 1/1  & $\emptyset$   & 1/1  \\
			Hindi (HI) & 1/1  & $\emptyset$   & 1/1  \\
			\bottomrule
		\end{tabular}
	}
	\caption{Annotated Data available per studied tasks for languages included in mT5 pretraining. The selection was done starting on the TyDiQA languages and extended to have a better coverage across all tasks. 1/1 means that data is available for both training and evaluation. $\emptyset$/1 means it is only available for evaluation. 
	}
	\label{tab:data_langs}
\end{table}

Each language is used both as a source language (fine-tuning) and as a target language (evaluation).

\subsection{Implementation}

Our experiments are based on the pre-trained Multilingual T5 models released by \cite{xue2020mt5} and available in the Transformers library \cite{wolf-etal-2020-transformers} based on which the fine-tuning experiments have been performed.

\subsection{Sequence Prediction}

With (m)T5, every NLP tasks is framed as a text generation task.
For QA and XNLI, we follow closely what has been introduced by \cite{raffel2019exploring}.
For NER, we use the gold tokenization provided in the dataset from \cite{ghaddar-langlais-2017-winer} and we perform sequence labeling by generating the sequence of labels along with the original input tokens as follows:

\begin{tabular}{r|l}
	\small
	Input & \textit{It \ was \ named \ for \ Williams \ College \ in \ Williamstown.}\\
	Output & It : O \ | \ was : O \ | \ named : O \ | \ for : O \ | \ Williams : B-ORG \ | \ College : I-ORG \\
	& in : O \ | \ Williamstown : B-LOC \ | \ . : O\\
\end{tabular}

\subsection{Hyperparameters}

\begin{table}[h!]
	\footnotesize 
	\centering
	\begin{tabular}{lrr}
		\toprule
		\textit{Params.}& QA  & Bounds\\
		\hline
		batch size & 128 &[1, 8192] \\
		Optimizer & Adam  & -\\
		learning rate &  3e-4 & [1e-6,1e-3] \\
		gradient clipping value &  1.0 & - \\
		Max Sequence Length (token)  & 512 & [1, 1024] \\
		\bottomrule
	\end{tabular}
	\caption{Fine-tuning best hyper-parameters. Reported best epoch indicates the best selected epoch when English is the source language in the zero-shot cross-lingual setting.}
	\label{tab:hyperparameters}
\end{table}

We fine-tune mT5 using the same set of hyperparameters for each task.
In contrast with the original implementation from \cite{raffel2019exploring}, we use the Adam Optimizer \cite{kingma2014adam}.
We report in Table~\ref{tab:hyperparameters} the set of hyperparameters used.

As highlighted by \cite{keung-etal-2020-dont}, cross-lingual performance can be unstable from one run to another.
To tackle this instability, each experiment is ran on at least 4 different random seeds.
Our analysis is computed on the average of the successful runs.

\section{Absolute Performance and Prediction of mT5 in the Cross-Lingual Setting}

We report in Table~\ref{fig:gold_squad_0}-\ref{fig:gold_XNLI_0} the zero-shot and few-shot (using $n=100$ as number of samples in the target language), for QA, NER and XNLI. 

\definecolor{mypink}{rgb}{0.5647058823529412, 0.9333333333333333, 0.5647058823529413}
\definecolor{inLang}{rgb}{0.5647058823529412,0.9333333333333333,0.5647058823529413} 
\definecolor{minus5}{rgb}{0.5716269171081199,0.92597656873066,0.6897434676489668} 
\definecolor{minus10}{rgb}{0.5785606125644822,0.9186071434268033,0.8052582998060297} 
\definecolor{minus15}{rgb}{0.5924659855807572,0.8000422023372783,0.903830310715539} 
\definecolor{minus20}{rgb}{0.6064220014017665,0.6064220014017665,0.8890028351995407} 
\definecolor{minus25}{rgb}{0.7028027023726197,0.6134190003640464,0.8815701063897662} 
\definecolor{above25}{rgb}{0.8666666666666667,0.6274509803921569,0.8666666666666667} 

\begin{figure*}
	\centering
	\begin{minipage}{0.45\textwidth}
		\centering
		\includegraphics[width=0.9\textwidth]{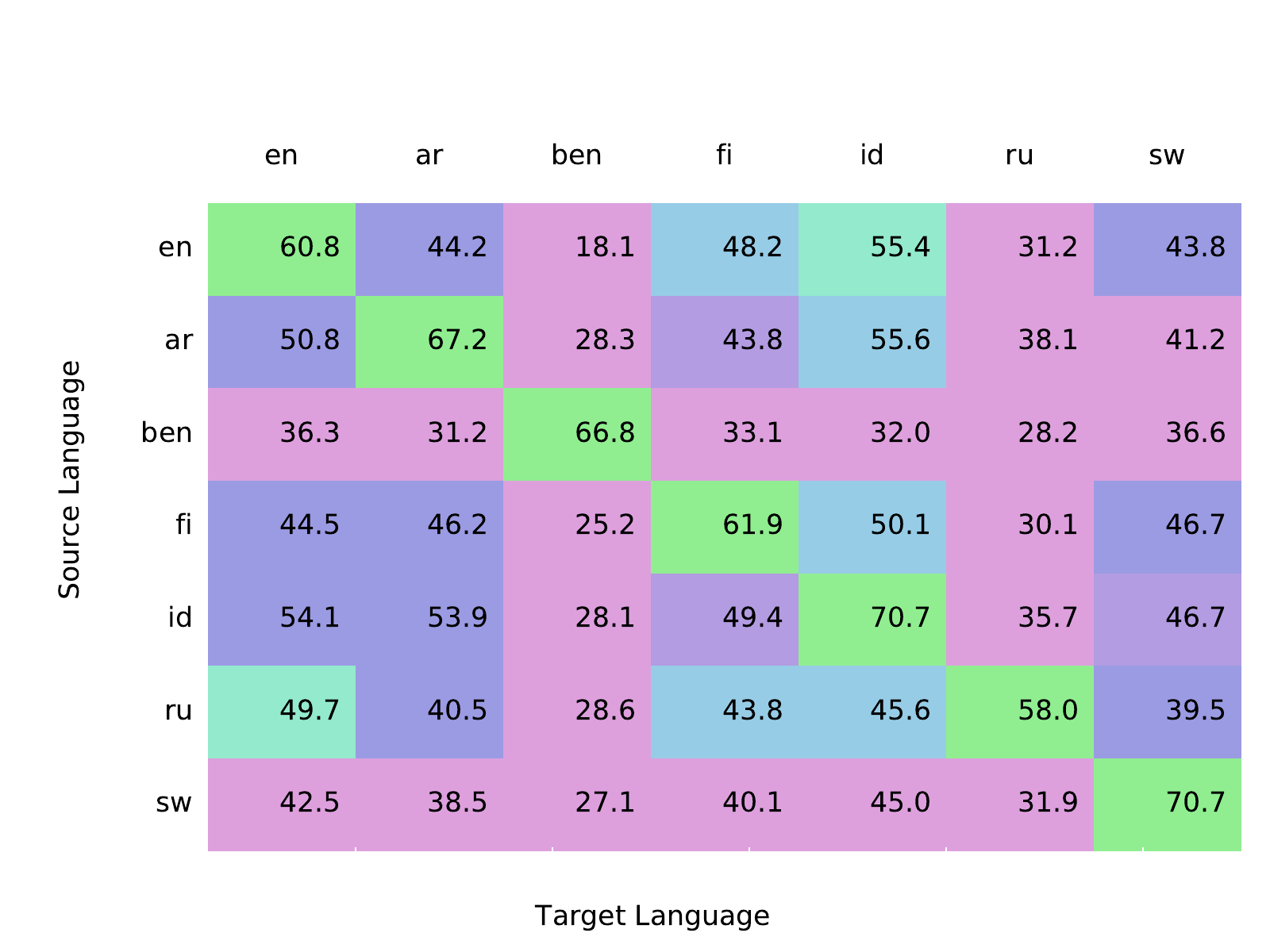}
		\subcaption{\small Observed Performance: coloring computed based on the cross-lingual gap which is equal to the cross-lingual performance on the target language subtracted from the performance on the source language.\\ \colorbox{inLang}{$\geq$ 0}
			\colorbox{minus5}{$\geq$ -5}
			\colorbox{minus10}{$\geq$ -10} 
			\colorbox{minus15}{$\geq$ -15} 
			\colorbox{minus20}{$\geq$ -20} 
			\colorbox{minus25}{$\geq$ -25} 
			\colorbox{above25}{$\leq$ -25}
		}
	\end{minipage}\hfill
	\begin{minipage}{0.45\textwidth}
		\centering
		\includegraphics[width=0.9\textwidth]{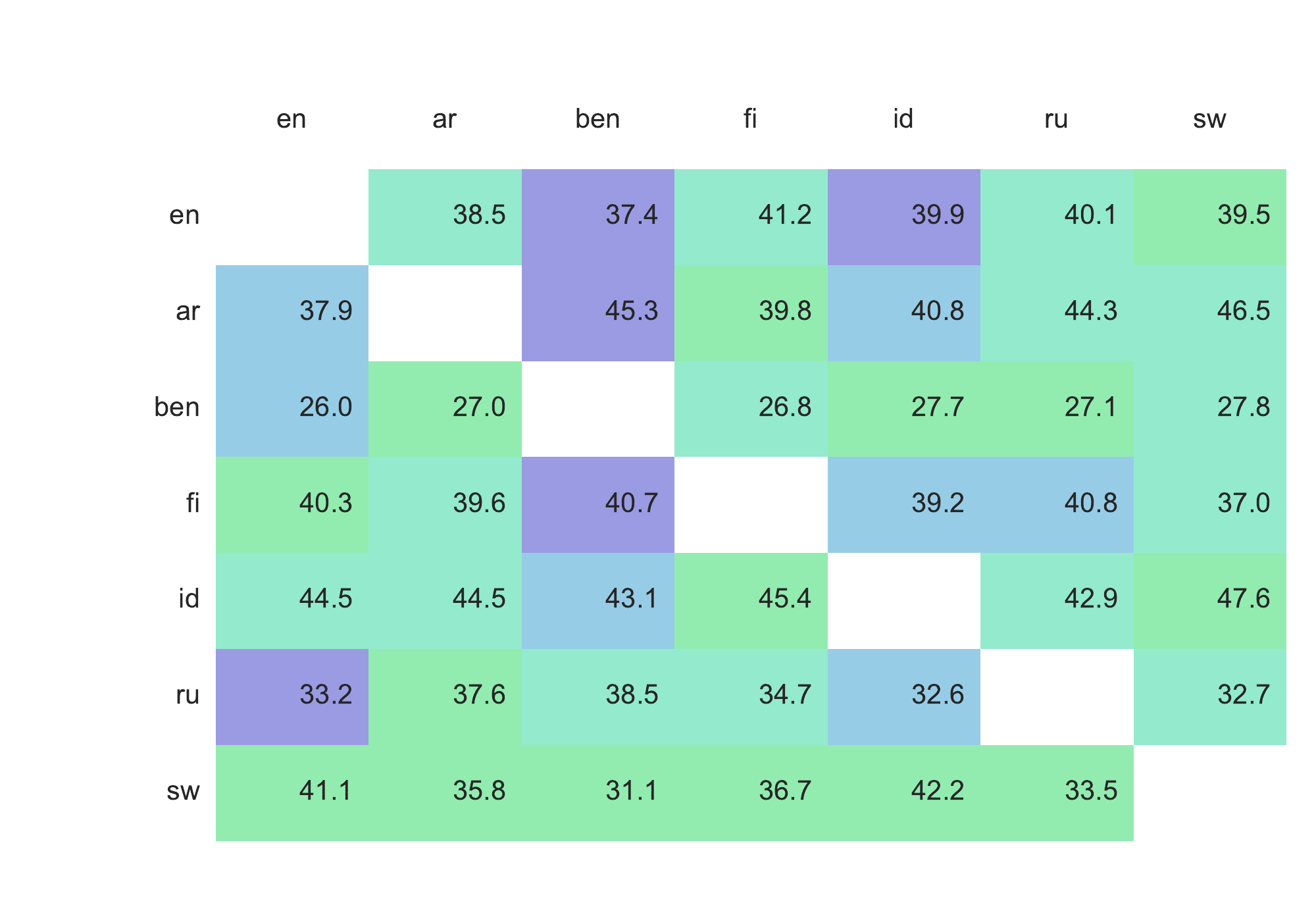}
		\subcaption{\small Predicted Performance: coloring computed based on the absolute prediction error (absolute difference between prediction and observed performance.\\
			\colorbox{inLang}{$=$ 0}
			\colorbox{minus5}{$\leq$ 5}
			\colorbox{minus10}{$\leq$ 10} 
			\colorbox{minus15}{$\leq$ 15}
			\colorbox{minus20}{$\leq$ 20} 
			\colorbox{minus25}{$\leq$ 25} 
			\colorbox{above25}{$\geq$ 25}
		}
	\end{minipage}
	\vspace{0.4cm}
	\caption{\textbf{Question Answering (QA)} \textbf{zero-shot} (0 annotated samples from the target language ($n=0$)) Exact Match Accuracy of mT5 base in the cross-lingual setting.}
	\label{fig:gold_squad_0}
\end{figure*}

\begin{figure*}
	\centering
	\begin{minipage}{0.45\textwidth}
		\centering
		\includegraphics[width=0.9\textwidth]{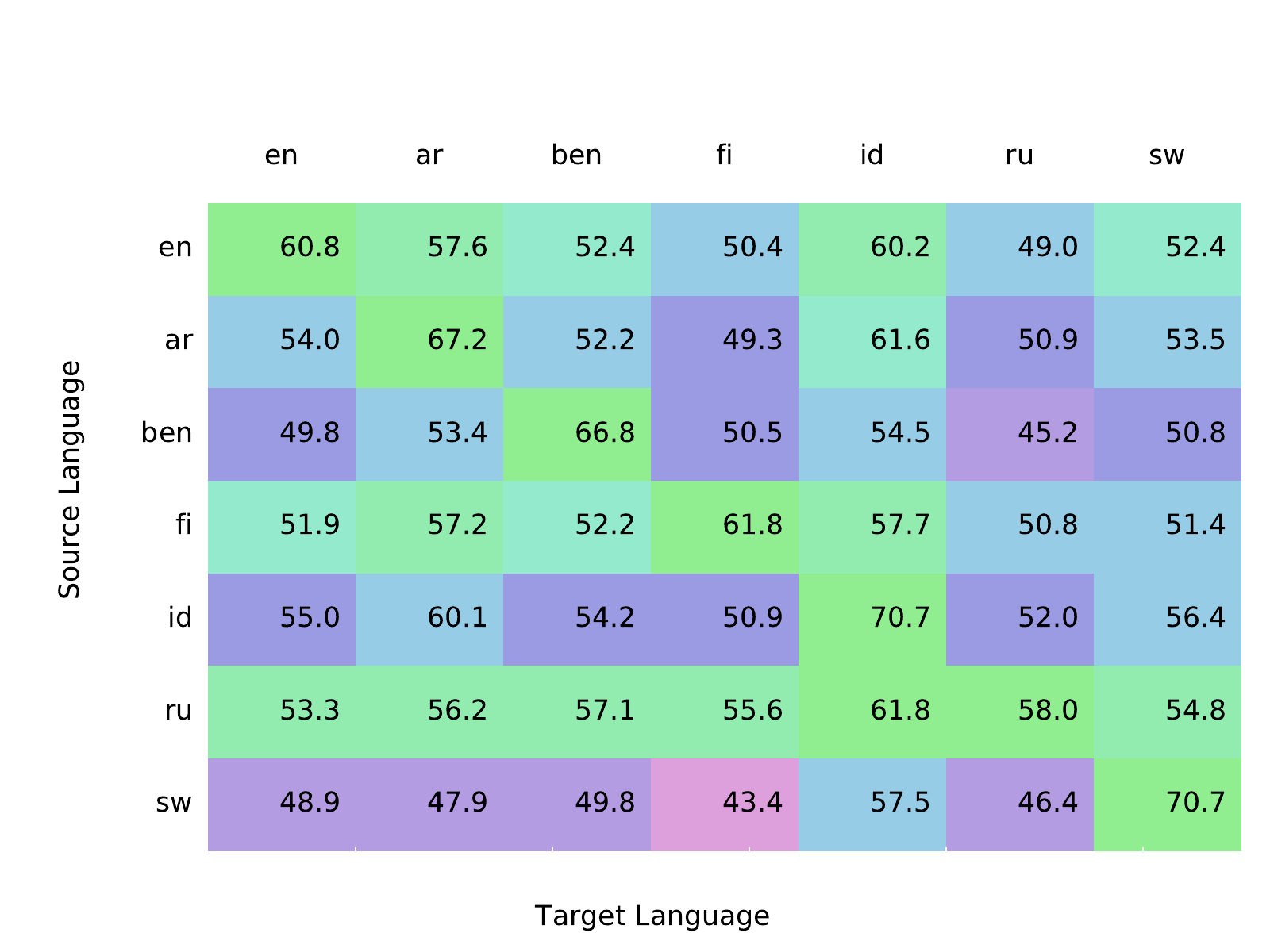}
		\subcaption{\small Observed Performance: coloring computed based on the cross-lingual gap which is equal to the cross-lingual performance on the target language subtracted from the performance on the source language.\\
			\colorbox{inLang}{$\geq$ 0}
			\colorbox{minus5}{$\geq$ -5}
			\colorbox{minus10}{$\geq$ -10}
			\colorbox{minus15}{$\geq$ -15}
			\colorbox{minus20}{$\geq$ -20}
			\colorbox{minus25}{$\geq$ -25}
			\colorbox{above25}{$\leq$ -25}
		}
	\end{minipage}\hfill
	\begin{minipage}{0.45\textwidth}
		\centering
		\includegraphics[width=0.9\textwidth]{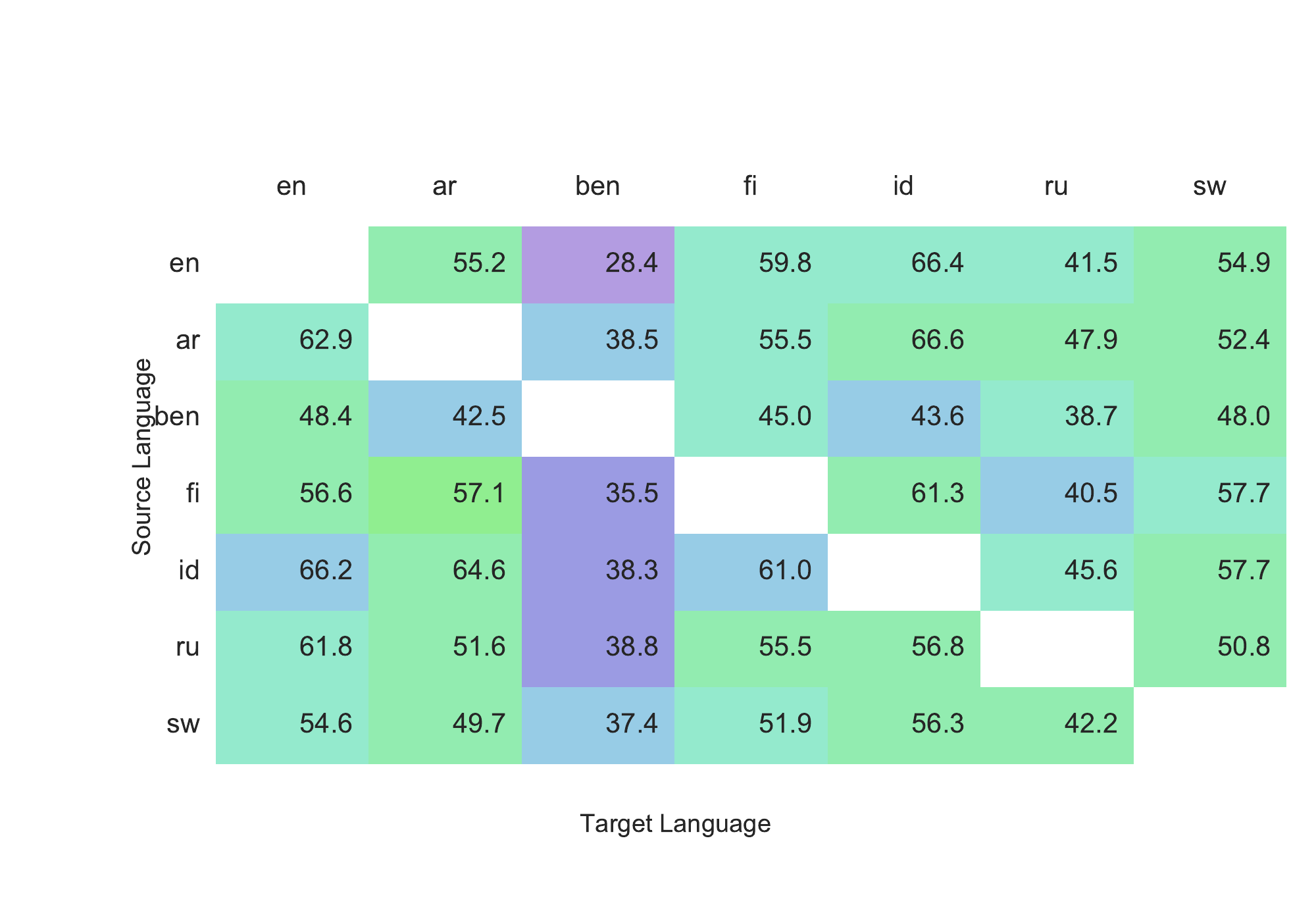}
		\subcaption{\small Predicted Performance: coloring computed based on the absolute prediction error (absolute difference between prediction and observed performance).\\
			\colorbox{inLang}{$=$ 0}
			\colorbox{minus5}{$\leq$ 5}
			\colorbox{minus10}{$\leq$ 10}
			\colorbox{minus15}{$\leq$ 15}
			\colorbox{minus20}{$\leq$ 20}
			\colorbox{minus25}{$\leq$ 25}
			\colorbox{above25}{$\geq$ 25}
		}
	\end{minipage}
	\vspace{0.4cm}
	\caption{\textbf{Question Answering (QA)} \textbf{few-shot} (100 annotated samples from the target language ($n=100$)) Exact Match Accuracy of mT5 base in the  cross-lingual setting.} 
\end{figure*}

\begin{figure*}
	\centering
	\begin{minipage}{0.45\textwidth}
		\centering
		\includegraphics[width=0.9\textwidth]{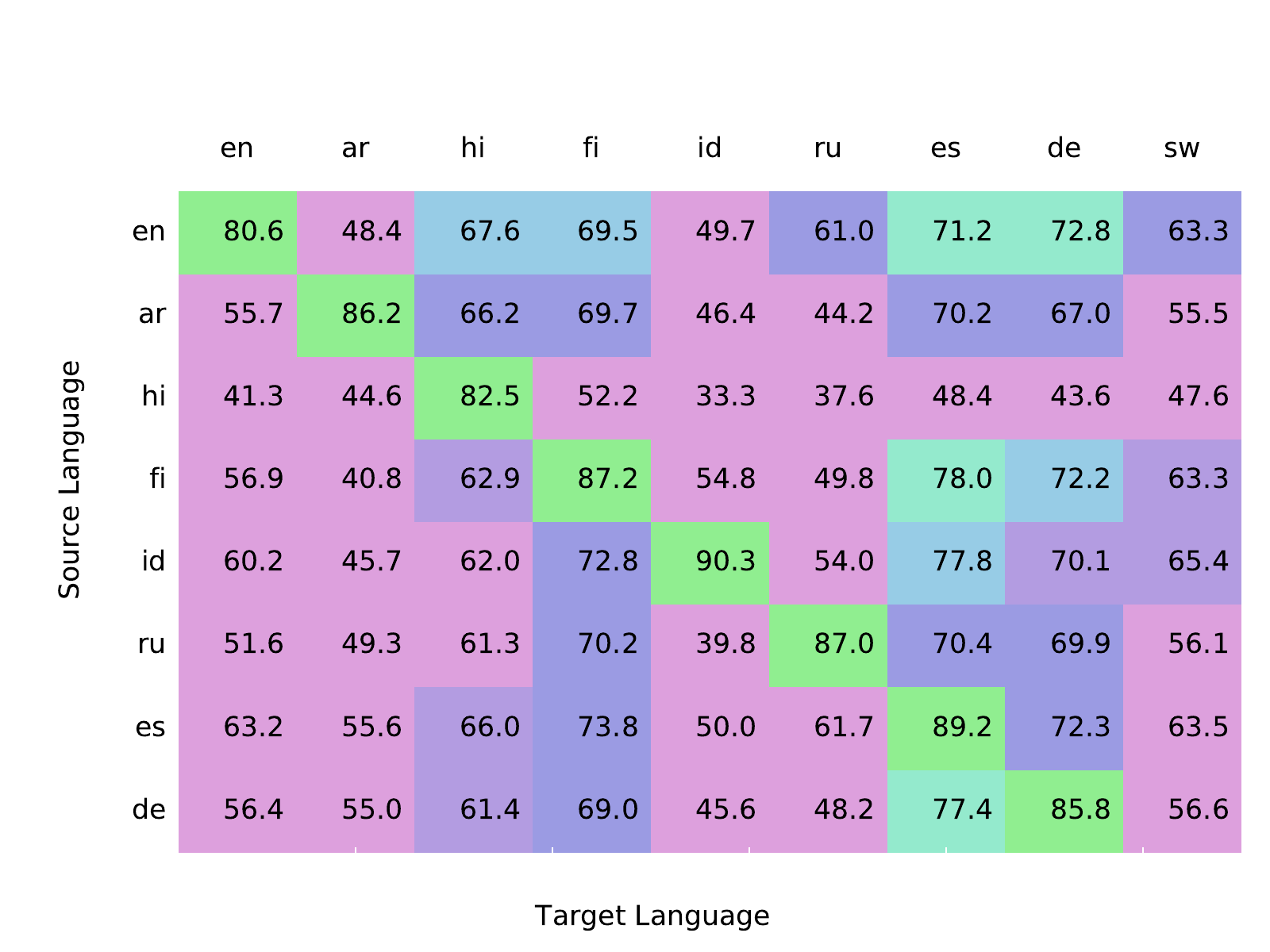}
		\subcaption{\small Observed Performance: coloring computed based on the cross-lingual gap which is equal to the cross-lingual performance on the target language subtracted from the performance on the source language.\\
			\colorbox{inLang}{$\geq$ 0}
			\colorbox{minus5}{$\geq$ -5}
			\colorbox{minus10}{$\geq$ -10}
			\colorbox{minus15}{$\geq$ -15}
			\colorbox{minus20}{$\geq$ -20}
			\colorbox{minus25}{$\geq$ -25}
			\colorbox{above25}{$\leq$ -25}
		}
	\end{minipage}\hfill
	\begin{minipage}{0.45\textwidth}
		\centering
		\includegraphics[width=0.9\textwidth]{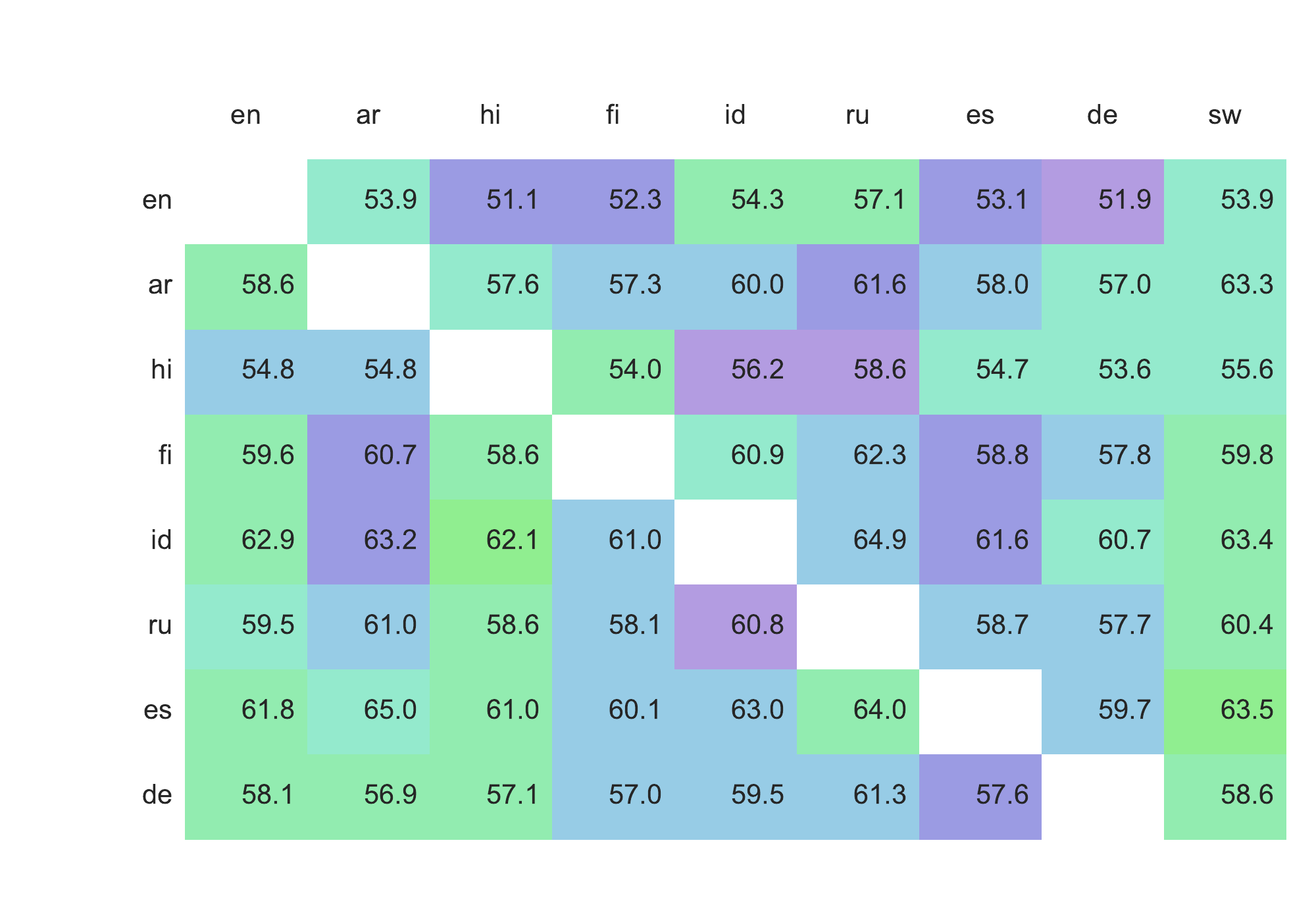}
		\subcaption{\small Predicted Performance: coloring computed based on the absolute prediction error (absolute difference between prediction and observed performance).\\
			\colorbox{inLang}{$=$ 0}
			\colorbox{minus5}{$\leq$ 5}
			\colorbox{minus10}{$\leq$ 10}
			\colorbox{minus15}{$\leq$ 15}
			\colorbox{minus20}{$\leq$ 20}
			\colorbox{minus25}{$\leq$ 25}
			\colorbox{above25}{$\geq$ 25}
		}
	\end{minipage}
	\vspace{0.4cm}
	\caption{\textbf{NER} \textbf{zero-shot} (0 annotated samples from the target language ($n=0$)) F1 score of mT5 base in the cross-lingual setting.} 
\end{figure*}

\begin{figure*}
	\centering
	\begin{minipage}{0.45\textwidth}
		\centering
		\includegraphics[width=0.9\textwidth]{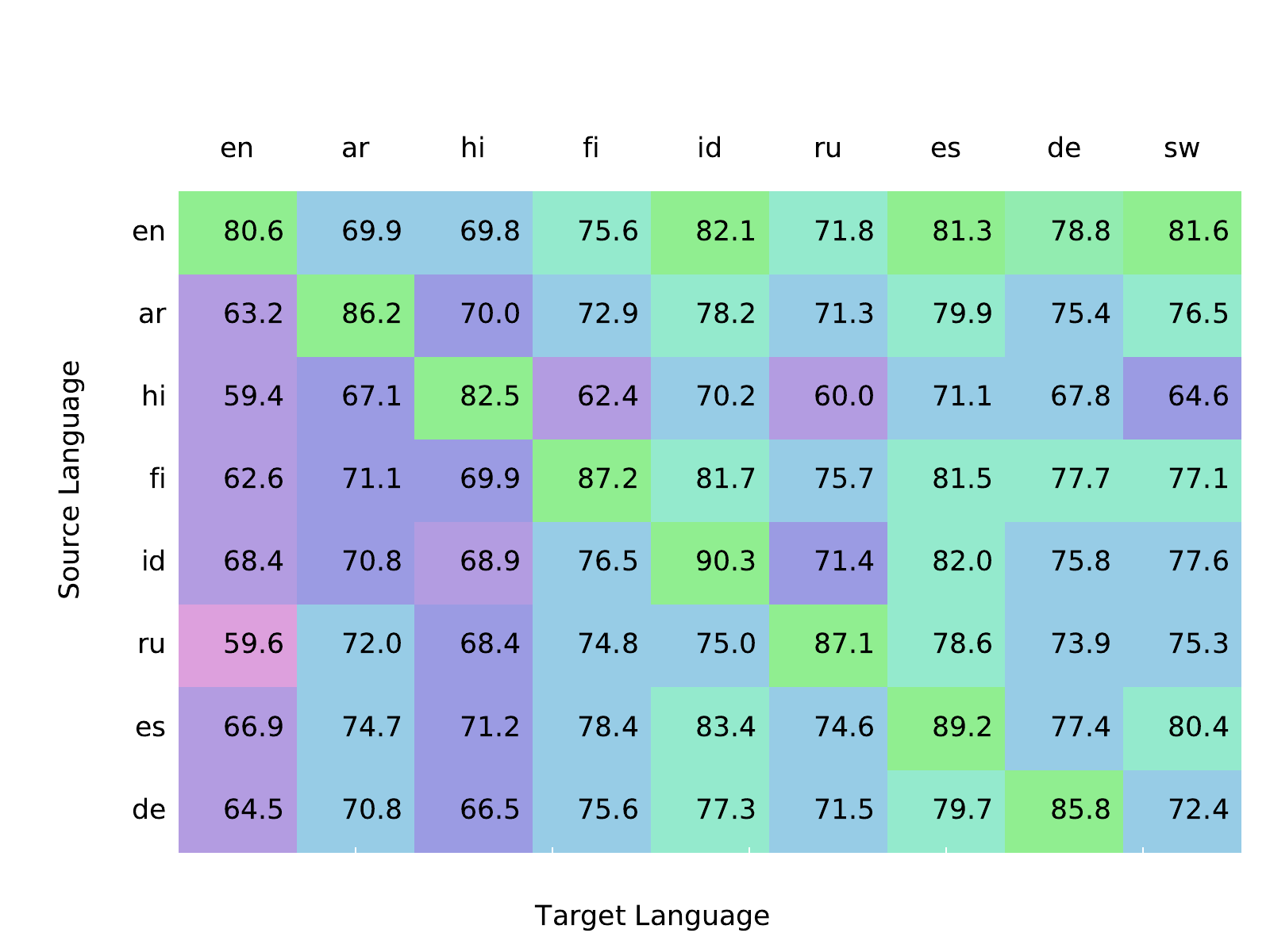}
		\subcaption{\small Observed Performance: coloring computed based on the cross-lingual gap which is equal to the cross-lingual performance on the target language subtracted from the performance on the source language.\\
			\colorbox{inLang}{$\geq$ 0}
			\colorbox{minus5}{$\geq$ -5}
			\colorbox{minus10}{$\geq$ -10}
			\colorbox{minus15}{$\geq$ -15}
			\colorbox{minus20}{$\geq$ -20}
			\colorbox{minus25}{$\geq$ -25}
			\colorbox{above25}{$\leq$ -25}
		}
	\end{minipage}\hfill
	\begin{minipage}{0.45\textwidth}
		\centering
		\includegraphics[width=0.9\textwidth]{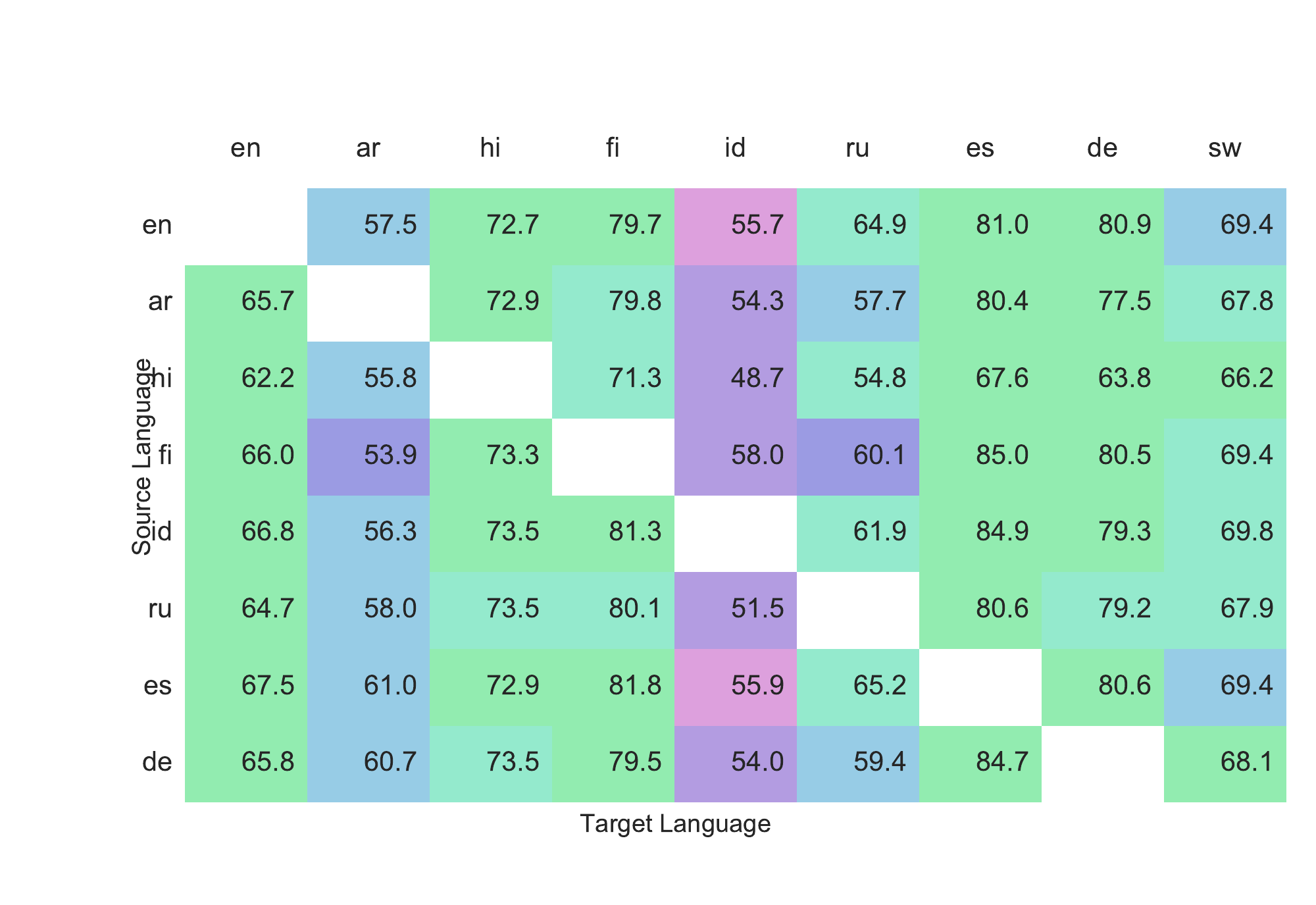}
		\subcaption{\small Predicted Performance: coloring computed based on the absolute prediction error (absolute difference between prediction and observed performance).\\
			\colorbox{inLang}{$=$ 0}
			\colorbox{minus5}{$\leq$ 5}
			\colorbox{minus10}{$\leq$ 10}
			\colorbox{minus15}{$\leq$ 15}
			\colorbox{minus20}{$\leq$ 20}
			\colorbox{minus25}{$\leq$ 25}
			\colorbox{above25}{$\geq$ 25}
		}
	\end{minipage}
	\vspace{0.4cm}
	\caption{\textbf{NER} \textbf{few-shot} (100 annotated samples from the target language ($n=100$)) F1 score of mT5 base in the  cross-lingual setting.} 
\end{figure*}

\begin{figure*}
	\centering
	\begin{minipage}{0.45\textwidth}
		\centering
		\includegraphics[width=0.9\textwidth]{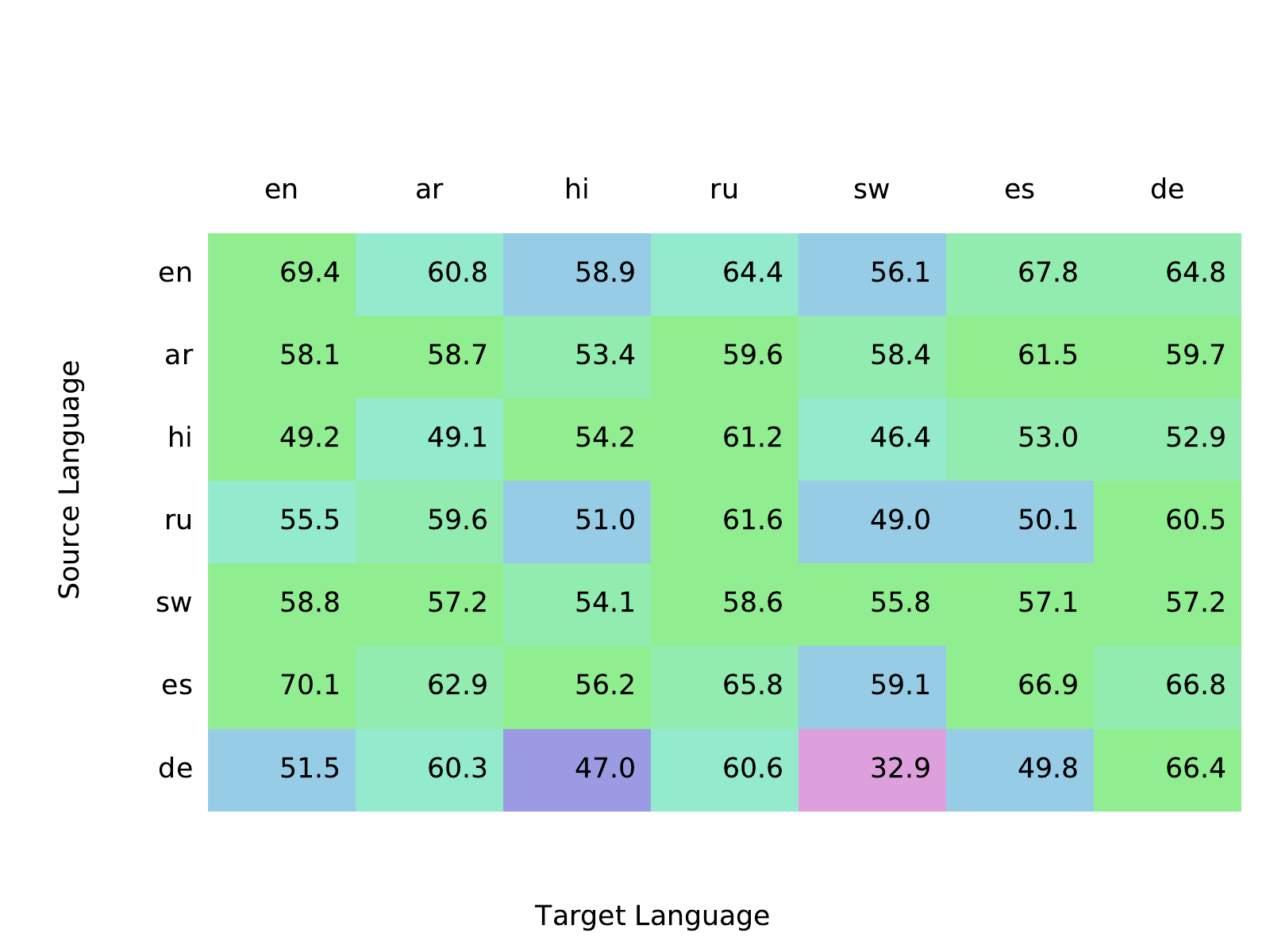}
		\subcaption{\small Observed Performance: coloring computed based on the cross-lingual gap which is equal to the cross-lingual performance on the target  language subtracted  from the performance on the source language.\\
			\colorbox{inLang}{$\geq$ 0}
			\colorbox{minus5}{$\geq$ -5}
			\colorbox{minus10}{$\geq$ -10}
			\colorbox{minus15}{$\geq$ -15}
			\colorbox{minus20}{$\geq$ -20}
			\colorbox{minus25}{$\geq$ -25}
			\colorbox{above25}{$\leq$ -25}
		}
	\end{minipage}\hfill
	\begin{minipage}{0.45\textwidth}
		\centering
		\includegraphics[width=0.9\textwidth]{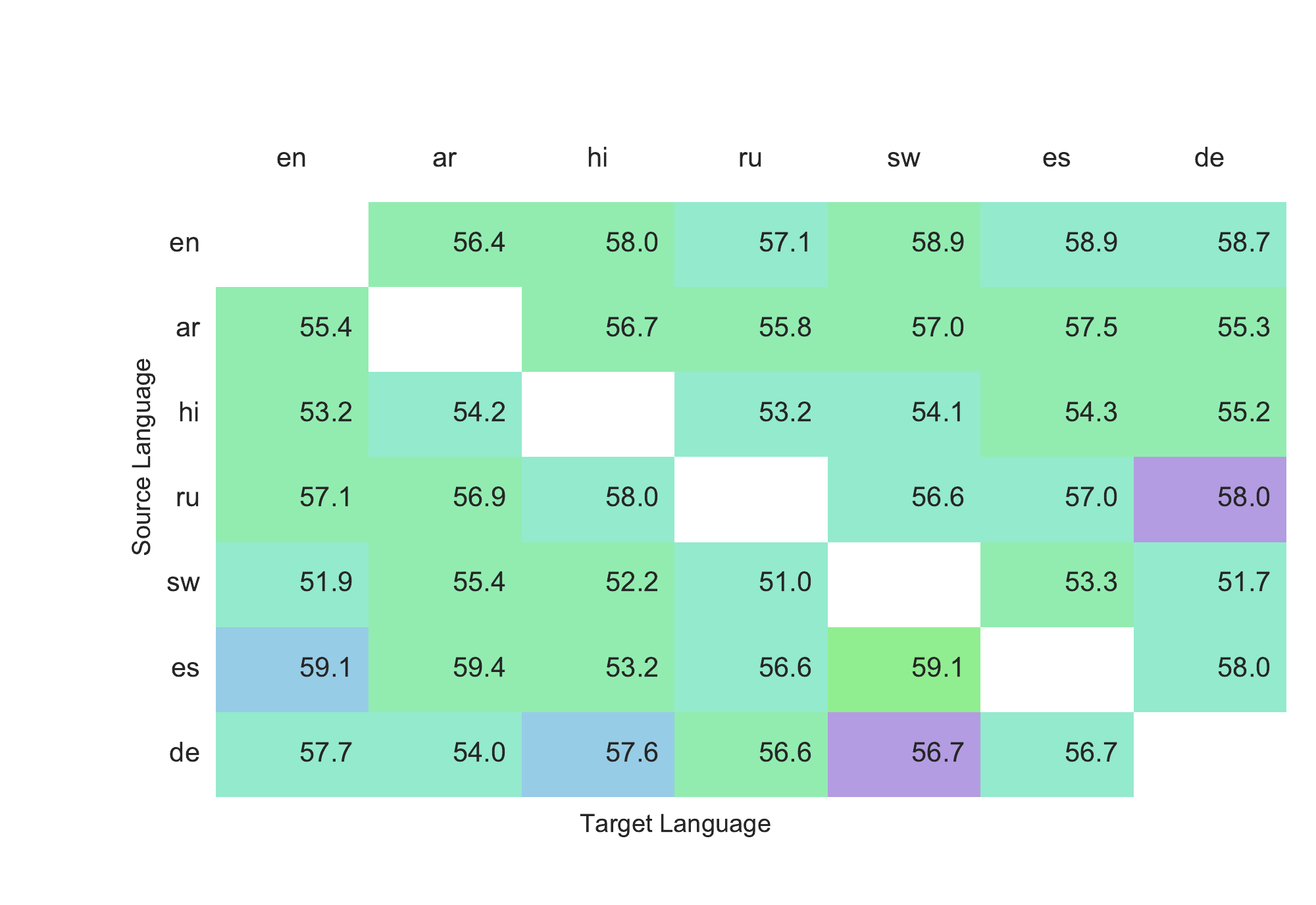}
		\subcaption{\small Predicted Performance: coloring computed based on the absolute prediction error (absolute difference between prediction and observed performance).\\ 
			\colorbox{minus5}{$\leq$ 5}
			\colorbox{minus10}{$\leq$ 10}
			\colorbox{minus15}{$\leq$ 15}
			\colorbox{minus20}{$\leq$ 20}
			\colorbox{minus25}{$\leq$ 25}
			\colorbox{above25}{$\geq$ 25}
		}
	\end{minipage}
	\vspace{0.4cm}
	\caption{\textbf{XNLI} \textbf{zero-shot} (0 annotated samples from the target language ($n=0$)) F1 score of mT5 base in the cross-lingual setting.} 
\end{figure*}

\begin{figure*}
	\centering
	\begin{minipage}{0.45\textwidth}
		\centering
		\includegraphics[width=0.9\textwidth]{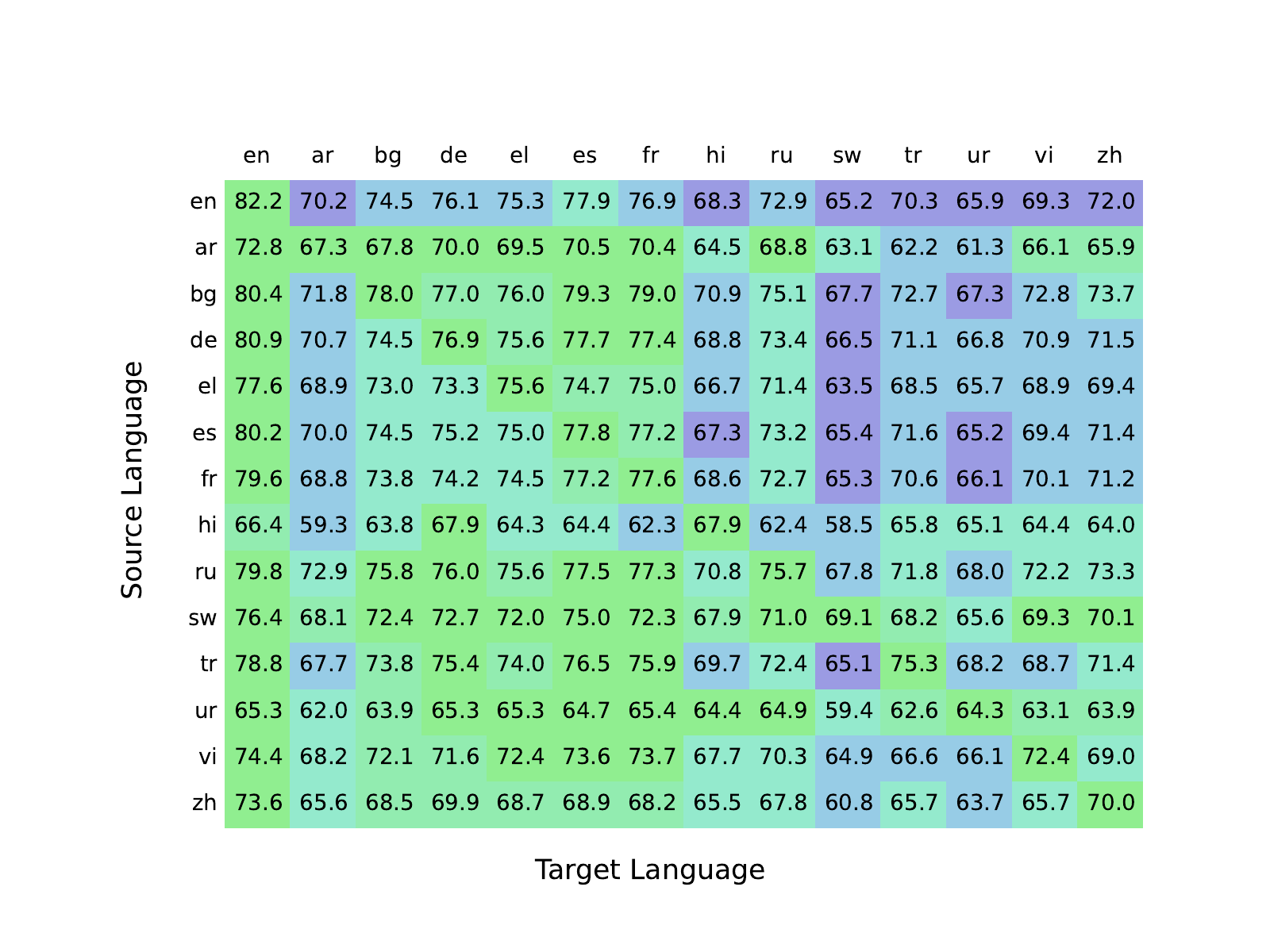}
		\subcaption{\small Observed Performance: coloring computed based on the cross-lingual gap which is equal to the cross-lingual performance on the target language subtracted from the performance on the source language.\\
			\colorbox{inLang}{$\geq$ 0}
			\colorbox{minus5}{$\geq$ -2}
			\colorbox{minus10}{$\geq$ -5}
			\colorbox{minus15}{$\geq$ -10}
			\colorbox{minus20}{$\geq$ -20}
			\colorbox{minus25}{$\geq$ -25}
			\colorbox{above25}{$\leq$ -25}
		}
	\end{minipage}\hfill
	\begin{minipage}{0.45\textwidth}
		\centering
		\includegraphics[width=0.9\textwidth]{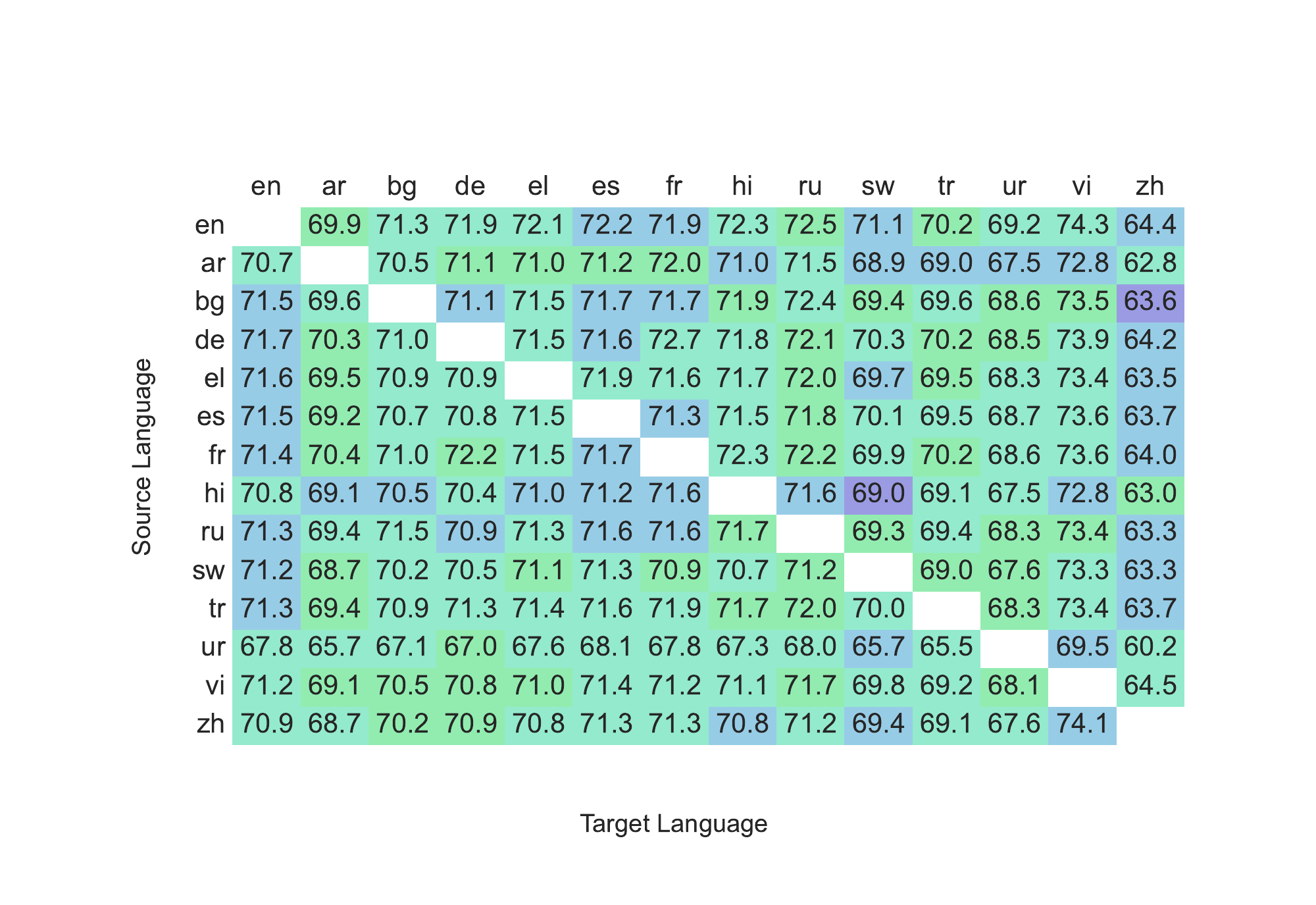}
		\subcaption{\small Predicted Performance: coloring computed based on the absolute prediction error (absolute difference between prediction and observed performance).\\ 
			\colorbox{minus5}{$\leq$ 2}
			\colorbox{minus10}{$\leq$ 5}
			\colorbox{minus15}{$\leq$ 10}
			\colorbox{minus20}{$\leq$ 20}
			\colorbox{minus25}{$\leq$ 25}
			\colorbox{above25}{$\geq$ 25}
		}
	\end{minipage}
	\vspace{0.4cm}
	\caption{\textbf{XNLI NEW} \textbf{zero-shot} (0 annotated samples from the target language ($n=0$)) F1 score of mT5 base in the  cross-lingual setting.} 
	\label{fig:gold_XNLI_0}
\end{figure*}


\end{document}